\documentclass[cameraready]{Interspeech}

\title{Evaluation of forced alignment of code-mixed speech: the case of Hindi-English}

\author[affiliation={1}, equalcontribution]{Ayushi}{Pandey}
\author[affiliation={1}, equalcontribution]{Pamir}{Gogoi}
\author[affiliation={2,3}, orcid=0000-0001-7382-9344, correspondingauthor]{Kevin}{Tang}

\address{
    $^1$ Karya, India \\
    $^2$ Department of English Language and Linguistics, Institute of English and American Studies, Faculty of Arts and Humanities, Heinrich Heine University Düsseldorf, Germany \\
    $^3$ Department of Linguistics, University of Florida, United States of America
}

\email{ayushi@karya.in, pamir.gogoi@karya.in, kevin.tang@hhu.de}

\keywords{code-mixing, code-switching, forced alignment, pronunciation variation, speech recognition}

\usepackage{comment}
\usepackage{multirow}
\usepackage[table]{xcolor}
\usepackage{graphicx}
\usepackage{float}
\usepackage{tipa}
\usepackage{polyglossia}
\setmainlanguage{english}
\setotherlanguage{hindi}

\newfontfamily\hindifont{NotoSansDevanagari.ttf}

\usepackage{hyperref}

\begin{document}

\maketitle

\begin{abstract}
Code-mixed speech poses unique challenges to forced alignment: expanded inventories, orthographic errors, and speaker variation. We evaluate forced alignment of Hindi-English code-mixed speech using the Montreal Forced Aligner. We address 2 problems: (1) free variation involving native vs non-native pairs and (2) phonemic boundary detection for mid-utterance English words. Bootstrapping strategies substantially outperform unmodified lexicons. Acoustic models trained on sentence-level code-mixed data achieve a mean error of 4.15ms, ie. ten times lower than monolingual Hindi (38.18ms) or isolated English (37.58ms) alternatives. Principled lexicon design and code-mixed training data are both essential for reliable alignment of bilingual speech.
\end{abstract}

\section{Introduction}
Code-mixing (word-level insertions of another language in a sentential frame) frequently occurs in quotidian speech of bilingual and multilingual communities. English, as one of the official languages of India, is recognized as a widespread medium of education, and enjoys a superior diglossic position within the country \cite{parasher1981indian}. Therefore, speakers of Indian languages exhibit routine switches between their regional mother tongues (L1) and English.  

The phenomenon of code-mixing has been actively addressed in the full stack of speech technologies for Indian languages: from dataset curation \cite{pandey2018phonetically, nayak2022l3cube, senthamizhselvi2025building}, speech recognition \cite{palivela2025code, bhogale2026towards}, to reasoning in language models, and in audio generation \cite{gourav2025code, murthy2025building} for voice interfaces. However, the foundational task of phonemic alignment in code-mixing has been largely overlooked for Indian languages. This means that the scope and understanding of computational resources required for phonological, linguistic analysis of code-mixing is significantly lacking.

Phonemic alignment, which produces time-aligned phoneme boundaries, is a critical component of large-scale phonological analysis of languages. Forced aligners \cite{mcauliffe17_interspeech, gorman2011prosodylab, bain2023whisperx} automate this process, and enable phoneticians to document low-resource languages \cite{chodroff2025comparing, wang2023evaulating, Tang_Bennett2019_bootstrapping_FA}, model speaker variation and complement their lab-produced results with real-world data \cite{chodroff2014burst}. Forced alignment of code-mixed speech offers a unique opportunity in studying the advanced phonemic inventory of bilingual speakers \cite{amazouz2019exploring, pandey2020understanding} and examine their results in real-world, often low-resource settings \cite{ahn-etal-2025-automatic}. Therefore, this paper conducts structured experiments for forced-alignment on code-mixed language, Hindi and English. It identifies, and addresses two problems that arise in such developments.

The first problem is of a phonological nature. \textbf{RQ1}: \textit{How can pronunciation variation in code-mixed speech be automatically modeled within a forced-alignment framework?}Bilingual production in Hindi frequently exhibits free variation in segments shared across languages. For example, the English word \textit{phone} may be realized as either [p\super{h}on] or [fon]. This is different from speaker variation of the same target phoneme, because the target sometimes shifts \textit{within} the speaker itself. The problem is further exacerbated when the script does not clearly specify the target phoneme. We address this issue in Experiment I, where we identify that the exact site of such variability exists in the \textipa{[\textdyoghlig
]} $\sim$ \textipa{[z]} variation. This issue has been addressed through bootstrapping methods. 

A second problem in code-mixed forced alignment is of the technical nature. \textbf{RQ2}:\textit{ What acoustic training data (code-mixed, monolingual Hindi, or in-corpus English) is most suitable for aligning mid-utterance English words in bilingual speech?}
Accurate phonemic boundary detection for mid-utterance English words depends on the composition of the acoustic training data. Acoustic models trained exclusively on monolingual Hindi or monolingual English may not adequately reflect bilingual production patterns. It therefore remains unclear whether sentence-level code-mixed training data provides advantages over monolingual alternatives for forced alignment \cite{ahn-etal-2025-automatic,ahn2025investigating}.

RQ1 addresses the problem of modeling bilingual segmental variability within the pronunciation lexicon, moving beyond monolingual phoneme assumptions to better reflect attested production patterns. RQ2 evaluates how the composition of acoustic training data influences alignment quality in mixed-language contexts, particularly for embedded English items. To address these questions, we conduct two experiments: Experiment 1 focuses on lexicon design for bilingual variability (RQ1), while Experiment 2 examines the effect of acoustic training data on alignment performance (RQ2). Together, these experiments provide a principled evaluation of lexicon design and acoustic model selection for forced alignment in code-mixed speech.

\section{Sources of variation: a problem for forced alignment}
Two types of variations are discussed here: the phonological and the orthographic. The phonological variation is an example of variation in the pronunciation of code-mixed speech, where the acoustic realization of the target phoneme is not fixed. The orthographic variation, on the other hand, is an example of variation in the written form. Even though the PBCM corpus comes from newspapers, some orthographic inconsistencies persist. A detailed explanation of both forms is written below:

\subsection{Segmental free variation}

\subsubsection{\textipa{[p\super h]} $\sim$ \textipa{[f]}}
Hindi's native voiceless bilabial aspirate [\textipa{p\super h}] alternates with the voiceless labiodental fricative [f]. Both sounds involve turbulent labiodental airflow, aspiration in the case of [\textipa{p\super h}], frication in the case of [f], which likely drives their perceptual overlap in bilingual speech. For example, the English word \textit{phone} may be realized as either [p\super{h}on] or [fon], and the Hindi word \textit{safal} (`successful') as either [sap\super{h}\textipa{@}l] or [saf\textipa{@}l]. There is a tendency in monolingual speakers to use the native phoneme \textipa{p\super h}], whereas a bilingual speaker may either alternate correctly, or default to the non-native phoneme [f]. 

\subsubsection{\textipa{[\textdyoghlig
]} $\sim$ \textipa{[z]}} In the case of loan words that contain the voiced alveolar fricative [z]  (e.g, zebra, zero), the Hindi's native voiced palatal affricate [\textipa{\textdyoghlig}] alternates with the voiced alveolar fricative [z]. For example, \textit{zebra} may be realized as either [\textipa{\textdyoghlig i:br@}] or [\textipa{zi:br@}]. 

There is a tendency in monolingual speakers to use the native phoneme, whereas a bilingual speaker can be expected to choose the correct alternative.. 
It is important to note here, that a similar pattern exists between loanwords from Urdu (e.g, zulfein (hair), fizool (waste)), where monolingual speakers default to their native counterparts described above. However, here we focus only on the code-mixing between English-Hindi.

\subsection{Orthographic variation}

Devanagari script employs a subscript diacritic, the \textit{nuqta} (\texthindi{\char"25CC\char"093C}),
to distinguish borrowed phonemes from their native 
counterparts: (i)  \texthindi{ज} [\textipa{\textdyoghlig}] vs.\ \texthindi{ज़} [z] and (ii) \texthindi{फ} [\textipa{p\super h}] vs.\ \texthindi{फ़} [f].

\noindent In written work, the nuqta is frequently omitted, neutralizing the orthographic contrast within each pair. This introduces
\textbf{systematic ambiguity} in grapheme-to-phoneme (G2P) conversion. The problem is further compounded by a cultural, native-speaker awareness of this omission. This means that \textbf{bilingual speakers} who know the diacritic is frequently omitted may \textbf{compensate}: when a [z]-word (like \textit{zebra}) is written with a bare \texthindi{ज},   they either infer that the nuqta was erroneously dropped and correctly produce the [z]. And when a \textdyoghlig-word (like \textit{juice}) is written without the nuqta, they remain faithful to the orthograph and produce a \textdyoghlig.  This yields \textbf{bidirectional variation}, in which both members of each pair surface in environments where only one was intended. Because of this inconsistencies, a one-to-one mapping between the grapheme to phoneme is rendered invalid.
\section{Data}

\subsection{Acoustic data}
The Phonetically Balanced Code Mixed (PBCM) corpus \cite{pandey2018phonetically} consists of 6,941\footnote{more phonetically balanced sentences were added after the original publication, which reports 6,126 utterances} phonetically balanced read-speech utterances recorded at IIIT-Hyderabad. Textual prompts for this corpora were sampled from selected sections (LifeStyle, Technology, Sports) of a leading national newspaper, Dainik Bhaskar. The dataset was recorded by 113 speakers (58 male, and 55 female), whose L1 was Hindi and who were educated in English medium schools. For the purpose of analysis, Hindi and English tags were manually assigned to each word-type in the corpus. 

\begin{table}[H]
\caption{Distributions of Hindi-English words and phonemes.}
\label{tab:corpus-stats}
\centering
\captionsetup{justification=centering}
\begin{tabular}{|c|c|c|}
\hline
\textbf{}                 & \textbf{Hindi} & \textbf{English} \\ \hline
word (types)     &     4,790           &    3,754              \\ \hline
word (tokens)    &    54,961            &   18,839               \\ \hline
phoneme (types)  &      73          &        52          \\ \hline
phoneme (tokens) &       194,672        &         97,137         \\ \hline
\end{tabular}
\end{table}

\subsection{Lexical Refinement and Phone-set Standardization}
To address the inherent limitations of off-the-shelf G2P systems for Hindi-English code-mixed data, we implemented a multi-stage refinement pipeline to generate a high-fidelity lexicon:

\begin{enumerate}
    \item \textbf{Bilingual Phoneme Mapping:} We harmonized the disparate phonetic outputs by mapping British English (\texttt{eng-uk}) G2P results onto a Hindi-proximate phonetic space. This ensured that English words in both Roman and Devanagri scripts shared a consistent acoustic representation.
    
    \item \textbf{Syllabic De-noising:} We manually addressed systematic G2P errors, specifically the over-insertion of schwas caused by incorrect syllabification rules in the \texttt{eSpeak} model.
    
    \item \textbf{Nasal Disambiguation:} To resolve the confusion between nasalized vowels and nasal consonants, we implemented phonologically conditioned nasal insertion (e.g., inserting /m/ before bilabials and /n/ before dentals).

    \item \textbf{Phonological Bootstrapping:} For ambiguous cases where orthography was under-specified (e.g., /p\super{h}/ vs /\textipa{f}/ and /\textipa{\textdyoghlig}/ vs /z/), we utilized bootstrapping to map voiced/aspirated segments to more robust voiceless proxies, significantly reducing alignment drift. More details are presented in Section 4.1.
\end{enumerate}
\subsection{Gold-standard pronunciation and alignment data}\label{subsec:gold}
For Experiment 1, 100 words for each of the free-variation categories were hand-annotated. Similarly for Experiment 2, 10\% of code-mixed English words were selected for creating gold-standard annotations. In other words, 6 utterances out of the ~62 utterances spoken by each speaker were randomly selected. Code-mixed words found in those utterances were first hand-annotated by one fluent Hindi speaker (second author), and cross-checked by another native Hindi speaker (first author). 

\section{Methods}
This section gives a detailed description of the experimental procedure for both the experiments. In the first experiment, we describe the handling of the pronunciation variants in the /\textipa{p\super h}-f/ and /\textipa{\textdyoghlig}-z/ context. In the second experiment, we describe the approach towards selecting the most suitable acoustic model for forced-alignment of English words. The code for these experiments is available\footnote{https://github.com/Ayushi113/mfa-hindi-code-mixed}.

\subsection{Experiment 1}\label{method:sec:exp1}
\textbf{Montreal Forced Aligner (MFA) v1.0} (a Kaldi-based GMM-HMM system using fMLLR) was employed throughout the experiments. Although the model is relatively less modern, its sufficiency of has been attested in recent works on Forced Alignment \cite{rousso2024tradition}. This architecture was tested with five bootstrapping configurations to identify the optimal proxy phones for two pairs of free variation:

\begin{enumerate}
    \item \textbf{Model 1 (No Mapping):} This serves as the control model where the lexicon remains unedited. Forced alignment is performed using raw G2P output, treating [\textipa{p\super h, f, \textdyoghlig, z}] as four distinct acoustic units.
    
    \item \textbf{Model 2 (Majority Baseline):} This model maps all instances of a variation to the single most frequent realization observed in the specific speaker's data, testing if a speaker-dependent dominant phone can override orthographic errors.
    
    \item \textbf{Model 3 (Dominant Baseline Mapping):} This version targets the primary native Hindi counterparts. It maps the aspirated [\textipa{p\super h}] to [\textipa{p}] (unaspirated) or the voiced affricate [\textipa{\textdyoghlig}] to [c] (voiceless), aiming to align the speech to the most structurally similar native phone.
    
    \item \textbf{Model 4 (Fricative/Sibilant Mapping):} This model maps variants to their closest fricative or sibilant substitutes (e.g., [f] or [z] to [s]), testing if the aligner performs better when targets are reduced to a generic [+strident] feature.
    
    \item \textbf{Model 5 (Maximum Bootstrapping):} The most aggressive reduction strategy, where both variants in a pair are mapped to two different proxy phones (e.g., \textipa{\textdyoghlig} to c and z to s). %
\end{enumerate}
\subsection{Experiment 2}\label{method:sec:exp2}

Three acoustic models were trained and were used to force-align the hand-annotated code-mixed English words from the corpus as described in Section~\ref{subsec:gold}. 

\begin{itemize}
    \item \textit{Full dataset:} All 6,941 sentences were used to train and align the sentence-level in the corpus. Acoustic-phonetic transcriptions for code-mixed English words were generated within the utterance. This experiment aimed to generate phonemic boundaries for code-mixed English words using naturally occurring Hindi-English sentence level data from the corpus.
    \item \textit{Hindi monolingual chunks:} Using the TextGridTools \cite{buschmeier2013textgridtools} package, we separated the audio data and TextGrids into contiguous Hindi phrase-level and English word-level chunks.  English words with Hindi inflections (for example: ``amerik-i") were excluded from the analysis. This sub-experiment aimed to evaluate the accuracy of monolingual acoustic data on aligning code-mixed English words from the same corpus. 
    \item \textit{English words}: Code-mixed English words extracted in the previous step were used to generate phoneme level alignment. This sub-experiment aimed to  examine the efficacy of word-level English data on aligning itself, with similar training and testing environments.
\end{itemize}
\begin{figure*}[t]
  \includegraphics[width=0.3\linewidth,, height=2.2cm]{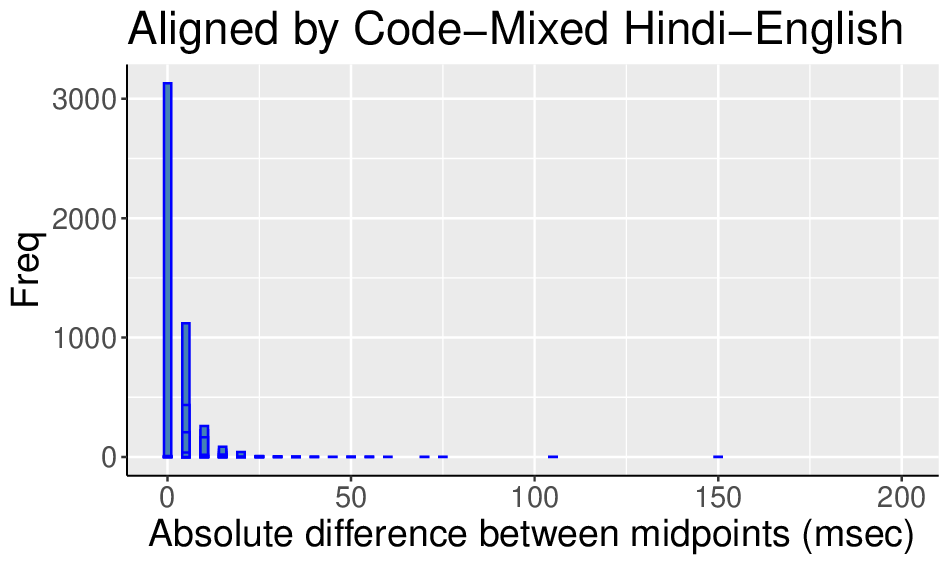}
  \includegraphics[width=0.3\linewidth,, height=2.2cm]{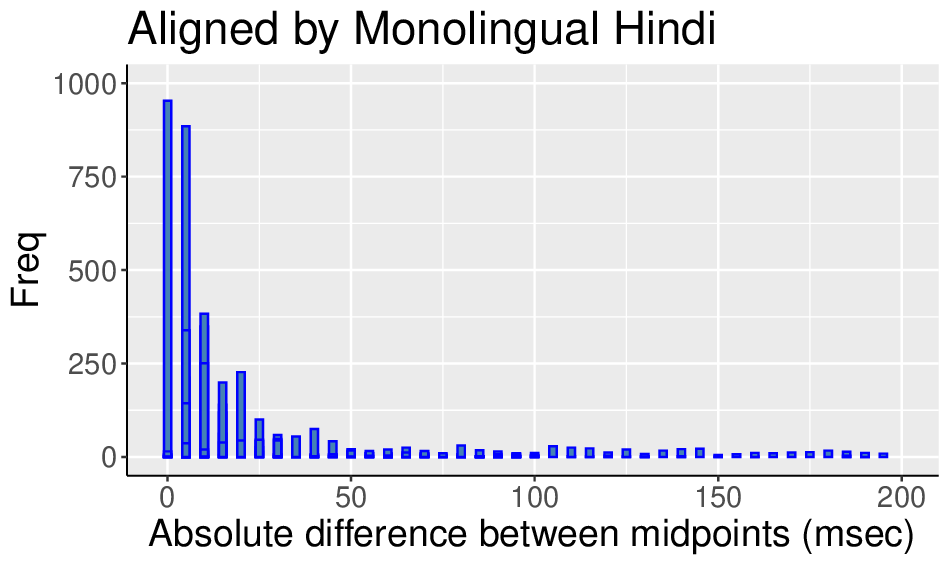}
  \includegraphics[width=0.3\linewidth, height=2.2cm]{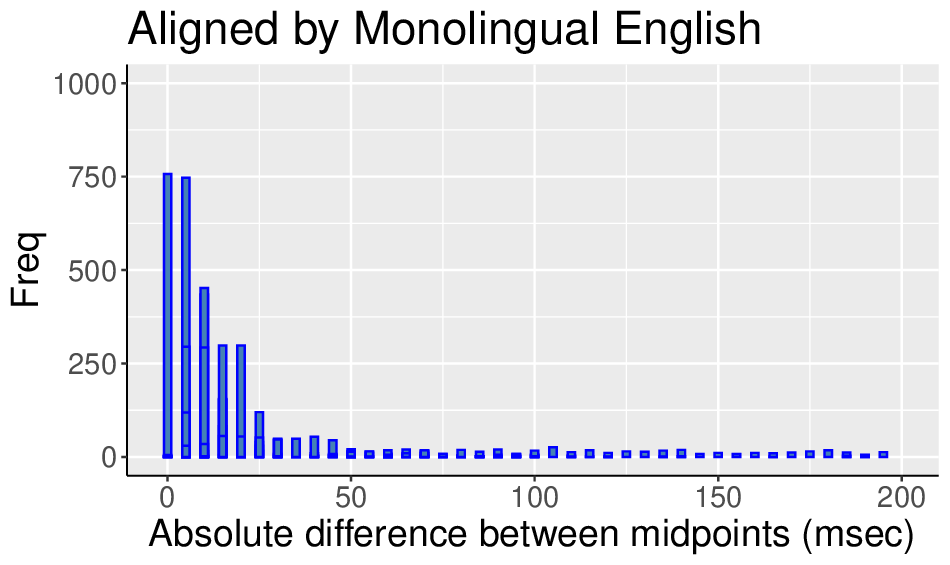}
  \caption{ Absolute difference between midpoints of phonemes in three models: Code-mixed (left),  Hindi (middle) and  English (right)}
  \label{fig:model_eval}
\end{figure*}

\section{Results and analysis}
In this section, we present the results obtained from the two experiments. The best performing model in Experiment 1 was chosen on the basis of the highest obtained F-score. For Experiment 2, the best training environment was chosen on the basis of absolute error in the midpoint values, when measured against the gold-standard annotations. 
\subsection{Experiment 1}\label{results:exp1}

\subsubsection{Words with the \textipa{[p\super h]} $\sim$ \textipa{[f]} variation}

\begin{table}[H]
\caption{Performance of five lexicon mapping strategies for resolving p\super h-f variation against a gold-standard transcript.
}
\label{tab:phwords_exp1}
\centering
\resizebox{\columnwidth}{!}{
\begin{tabular}{cccccc}
\toprule
\multicolumn{1}{l}{} & \textbf{\begin{tabular}[c]{@{}c@{}}No \\ mapping\end{tabular}} & \textbf{\begin{tabular}[c]{@{}c@{}}Majority \\ baseline\end{tabular}} & \textipa{p\super h}→\textipa{p} & f→s & \begin{tabular}[c]{@{}c@{}}\textipa{p\super h}→\textipa{p}\\ f→s\end{tabular} \\ \hline
\textbf{Accuracy}    & 0.56                                                           & 1                                                                     & 0.95            & 0.51           & 0.73                                                        \\
\textbf{Precision}   & 1                                                              & 1                                                                     & 1               & 0.1            & 1                                                           \\
\textbf{Recall}      & 0.56                                                           & 1                                                                     & 0.95            & 0.51           & 0.73                                                        \\
\textbf{F-score}     & 0.72                                                           & 1                                                                     & 0.97            & 0.37           & 0.84                                                        \\ \hline
\bottomrule
\end{tabular}}
\end{table}
Table 2 describes the diagonal accuracies of the confusion matrix between the gold-standard phoneme labels, and those predicted by different variations of the MFA. 
The results in Table \ref{tab:phwords_exp1} reveal a  discrepancy between orthographic representation and acoustic reality in Hindi-English code-mixed speech. While the No mapping column shows an F-score of only 0.72, the gold-standard data reveals that the speakers in this demographic almost unanimously produced the fricative /\textipa{f}/, despite the ``nukta-less" Hindi script suggesting the aspirated stop /\textipa{p\super h}/. This mismatch causes the MFA to trigger frequent alignment errors when forced to use the \texttt{eSpeak} default. Interestingly, the data suggests that for these specific code-mixed terms, the expected ``free variation" is absent in actual production; the speakers in the PBCM corpus have stabilized on the /\textipa{f}/ variant.

As seen in the \textipa{p\super h} → \textipa{p} column, simply removing aspiration provides a massive boost to an F-score of 0.97. This works because reverting to a plain labio-dental or bilabial stop creates a sufficiently distinct acoustic category from the fricative /\textipa{f}/, allowing the aligner to better distinguish the segments. In contrast, the f → s mapping fails , as the sibilant /s/ is acoustically too distant from the target. Given that the Majority baseline hits a perfect 1.0, the most robust recommendation from this data is that for code-mixed Hindi-English corpora, a direct lexicon mapping to /\textipa{f}/, or at minimum, de-aspirating to /\textipa{p}/, is necessary to correct for the orthographic ``nukta" deficit.

\subsubsection{Words with the \textipa{[\textdyoghlig
]} $\sim$ \textipa{[z]} variation}

\begin{table}[h]
\caption{Performance of five lexicon mapping strategies for resolving \textipa{\textdyoghlig}-z variation against a gold-standard transcript.}

\label{tab:jzwords_exp1}
\centering
\resizebox{\columnwidth}{!}{
\begin{tabular}{@{}cccccc@{}}
\toprule
\multicolumn{1}{l}{} & \textbf{\begin{tabular}[c]{@{}c@{}}No \\ mapping\end{tabular}} & \textbf{\begin{tabular}[c]{@{}c@{}}Majority \\ baseline\end{tabular}} & \textipa{\textdyoghlig}→c & z→s & \begin{tabular}[c]{@{}c@{}}\textipa{\textdyoghlig}→c\\ z→s\end{tabular} \\ \hline
\textbf{Accuracy}    & 0.16                & 0.69                          & 0.59            & 0.73           & 0.78                                                        \\
\textbf{Precision}   & 0.22                & 0.47                          & 0.79            & 0.75           & 0.84                                                           \\
\textbf{Recall}      & 0.16                & 0.69                          & 0.59            & 0.73           & 0.78                                                        \\
\textbf{F-score}     & 0.16                & 0.56                          & 0.59            & 0.66           & 0.74                                                        \\ \hline
\bottomrule
\end{tabular}}
\end{table}

The results in Table \ref{tab:jzwords_exp1} for the \textipa{\textdyoghlig}-z variation reflect a more complex alignment task than the previous set, as the ground truth contains both the voiced affricate and the voiced sibilant. The No mapping baseline fails significantly with an F-score of 0.16, indicating that the default G2P's singular output is incompatible with the speakers' actual productions. Even the Majority baseline, which defaults to the most frequent variant in the gold standard, only achieves an F-score of 0.56, highlighting that a single-label approach cannot sufficiently capture the variation present in words like ``Rajput'' or ``Resistance''.

As shown in the table, providing the aligner with multiple pronunciation variants, allowing a ``competition" between labels, yields a substantial performance increase. While the individual mappings \textipa{\textdyoghlig} → c and z → s show incremental gains, the maximum bootstrapping approach (\textipa{\textdyoghlig} → c and z → s) proves most effective with an F-score of 0.74 and a Precision of 0.84. These results suggest that by mapping both voiced targets to their voiceless counterparts in the lexicon, the MFA can more reliably identify the phoneme boundaries and recover the intended labels from the acoustic signal.

\subsection{Experiment 2}
The results of Experiment 1 gave us a variant-free lexicon, where different pronunciation variants were encoded as different lexical items. This design reduced ambiguity and enabled a clearer investigation of RQ2 by preventing free-variation errors from propagating to the subsequent experiment.
As discussed in Section~\ref{method:sec:exp2}, there were three different types of acoustic models for forced alignment. For evaluation of the obtained TextGrids, we had 4 datasets: a) the hand-annotated boundaries, and forced-aligned boundaries from b) the code-mixed sentence level data, c) the phrase-level monolingual Hindi chunks, and d) the word-level English only chunks. For every phoneme, the timestamp of the \textit{midpoint} of the left and right boundaries was extracted, for each of the forced-alignment conditions. Then, the midpoint was compared against the gold-standard midpoint, and absolute errors were computed. Figure~\ref{fig:model_eval} displays the comparative distribution of the absolute errors for each of the forced-aligned conditions. We observe that the code-mixed sentence level alignment outperform the two monolingual conditions. 

The mean absolute error by the code-mixed sentence level model was 4.15 ms, which is 10 times lower than the monolingual Hindi (38.18 ms) and English models (37.58 ms).

\begin{table}
\caption{Comparison of tolerance (in msec) of the three models}
    \label{tab:eval}
\resizebox{\columnwidth}{!}{
\begin{tabular}{llllll}
\toprule
\multirow{2}{*}{}             & \multicolumn{5}{c}{\textbf{Tolerance (msec)}}                              \\
                              & \textbf{\textless10} & \textbf{\textless20} & \textbf{\textless30} & \textbf{\textless40} & \textbf{\textless50} \\ \hline
\textbf{Code-mixed sentences} & 87.06       & 98.26       & 99.47       & 99.66       & 99.78       \\
\textbf{Mono Hindi chunks}    & 48.10       & 68.71       & 77.81       & 81.34       & 83.76       \\
\textbf{Mono English words}   & 41.89       & 68.29       & 79.05       & 82.58       & 84.82 \\ \hline
\bottomrule
\end{tabular}}
\end{table}

Table~\ref{tab:eval} displays the coverage of phonemes under each training environment, for different levels of error tolerance. For example, when trained with code-mixed sentences, 87.06\% phonemes were found to comparable with the gold-standard annotations, with an error of \textless 10 ms. As can clearly be seen (in a row-wise/model comparison), the model trained on code-mixed sentences shows the highest percentage of phonemes in the lower error (\textless 10 ms) range. A closer look reveals that when monolingual Hindi chunks are compared with word-level English as training, the former outperforms the latter only at low tolerance (\textless 10/20), but then the pattern reverses from 30ms onwards. This indicates that while on average, many tokens of Hindi models were better aligned (e.g. 7\% of the phones were better aligned at 10\% tolerance), it nonetheless contains more tokens that are particularly poorly aligned (\textgreater 30ms errors). In other words, the word-level English model has on average 1-2\% fewer tokens with errors of \textgreater 30ms, than the monolingual Hindi model.

\section{Discussion \& conclusion}
In this paper, we addressed the problem of free variation, and orthographic inconsistencies in code-mixed Hindi-English as a challenge for forced alignment. We found that bootstrapping techniques, used commonly in low-resource languages are useful for addressing such variation. Next, we identified that the use of an acoustic model trained on code-mixed sentences is most suitable for accurate alignment of the test corpus in code-mixed speech. An important observation is that code-mixed data, even at the word-level is more useful than a larger monolingual corpus of Hindi. 

Our work is among the first to present a structured analysis of computational tools for a phonological analysis of code-mixing in Indian languages, and extends previous research \cite{pandey2020understanding, ahn-etal-2025-automatic, amazouz2019exploring}. We have shown that speakers compensate for the inconsistency in orthography caused by the missing diacritic ("nukta"). In the ph-f variation, most speakers of our dataset defaulted to the /f/, and did not use the aspirated stop. The stop is more characteristic of a vernacular expression, and it is possible that in read speech, its manifestation is excluded. However, the \textipa{\textdyoghlig} words present a unique challenge where speaker compensation does not lead to a single majority variant. In words where the script omits the nukta, speakers exercise their linguistic competence to produce both \textipa{\textdyoghlig} and \textipa{z} depending on the intended lexical target. This necessitates the use of maximum bootstrapping (\textipa{\textdyoghlig} → c and z → s), which provides the aligner with the phonetic flexibility to recover both variants where a simple lexicon correction cannot. This confirms that while these speakers possess stable phonemic categories, the ``invisible" nukta in the script triggers a split in production. This problem is addressed in this paper through bootstrapping.

\section{Limitation}
At the time the experiments were conducted, an older version of Montreal Forced Aligner (MFA) was used. Repeating the experiments with a recent version of MFA (e.g., v3.x) may provide additional insights. In Experiment 1, pronunciation probabilities could be explicitly defined and conditioned on speaker demographics (e.g., education level), and pretrained G2P models for Hindi, Indian English, and British English could be incorporated. In Experiment 2, future work could experiment with the pretrained English acoustic model (v3.1.0), trained on multiple English varieties, including Indian and British English.
\section{Generative AI Use Disclosure}
Generative AI was used only for editing and polishing manuscripts, and not for producing a significant part of the manuscript.

\bibliographystyle{IEEEtran}
\bibliography{mybib}

@inproceedings{mcauliffe17_interspeech,
  author={McAuliffe, Michael and Socolof, Michaela and Mihuc, Sarah and Wagner, Michael and Sonderegger, Morgan},
  title={{Montreal Forced Aligner: Trainable Text-Speech Alignment Using Kaldi}},
  year=2017,
  booktitle={Proc. Interspeech 2017},
  pages={498--502},
  doi={10.21437/Interspeech.2017-1386},
  URL = {https://doi.org/10.21437/Interspeech.2017-1386}
}

@inproceedings{nayak2022l3cube,
  title={L3Cube-HingCorpus and HingBERT: A code mixed Hindi-English dataset and BERT language models},
  author={Nayak, Ravindra and Joshi, Raviraj},
  booktitle={Proceedings of the WILDRE-6 Workshop within the 13th Language Resources and Evaluation Conference},
  pages={7--12},
  year={2022}
}

@inproceedings{senthamizhselvi2025building,
  title={Building Code-Mixed Datasets for Multilingual Data Analysis: Methods, Metrics and Challenges},
  author={Senthamizhselvi, S and Chitrakala, S},
  booktitle={International Conference on Data Analytics \& Management},
  pages={564--574},
  year={2025},
  organization={Springer}
}

@article{rousso2024tradition,
  title={Tradition or innovation: A comparison of modern ASR methods for forced alignment},
  author={Rousso, Rotem and Cohen, Eyal and Keshet, Joseph and Chodroff, Eleanor},
  journal={Proceedings of InterSpeech 2024},
  pages={1525--1530},
  year={2024}
}

@article{palivela2025code,
  title={Code-switching ASR for low-resource Indic languages: A Hindi-Marathi case study},
  author={Palivela, Hemant and Narvekar, Meera and Asirvatham, David and Bhushan, Shashi and Rishiwal, Vinay and Agarwal, Udit},
  journal={IEEE Access},
  volume={13},
  pages={9171--9198},
  year={2025},
  publisher={IEEE}
}

@article{gorman2011prosodylab,
  title={Prosodylab-aligner: A tool for forced alignment of laboratory speech},
  author={Gorman, Kyle and Howell, Jonathan and Wagner, Michael},
  journal={Canadian acoustics},
  volume={39},
  number={3},
  pages={192--193},
  year={2011}
}

@inproceedings{wang2023evaulating,
  title={Evaulating forced alignment for under-resourced languages: A test on Squliq Atayal data},
  author={Wang, Chi-Wei and Chen, Bo-Wei and Huang, Po-Hsuan and Lai, Ching-Hung and Chiu, Chenhao},
  booktitle={Proceedings of the 20th International Congress of Phonetic Sciences},
  pages={3355--3359},
  year={2023},
  organization={Guarant International Prague, Czechia}
}

@article{chodroff2014burst,
  title={Burst spectrum as a cue for the stop voicing contrast in American English},
  author={Chodroff, Eleanor and Wilson, Colin},
  journal={The Journal of the Acoustical Society of America},
  volume={136},
  number={5},
  pages={2762--2772},
  year={2014},
  publisher={AIP Publishing}
}

@article{chodroff2025comparing,
  title={Comparing language-specific and cross-language acoustic models for low-resource phonetic forced alignment},
  author={Chodroff, Eleanor and Ahn, Emily P and Dolatian, Hossep},
  year={2025},
  publisher={University of Hawaii Press}
}

@inproceedings{murthy2025building,
  title={Building zero shot and code-mixed/switched synthesis systems for subcontinents with a rich language diversity},
  author={Murthy, Hema A},
  booktitle={Proc. SSW 2025},
  year={2025}
}

@article{bain2023whisperx,
  title={Whisperx: Time-accurate speech transcription of long-form audio},
  author={Bain, Max and Huh, Jaesung and Han, Tengda and Zisserman, Andrew},
  pages={4489--4494},
  journal={Proceedings of InterSpeech},
  year={2023}
}

@inproceedings{pandey2020understanding,
  title={Understanding forced alignment errors in Hindi-English code-mixed speech--a feature analysis},
  author={Pandey, Ayushi and Gogoi, Pamir and Tang, Kevin},
  booktitle={Proceedings of the First Workshop on Speech Technologies for Code-Switching in Multilingual Communities},
  pages={13--17},
  year={2020}
}

@inproceedings{amazouz2019exploring,
  title={Exploring consonantal variation in French-Arabic Code switching speech: the case of gemination},
  author={Amazouz, Djegdjiga and Adda-Decker, Martine and Lamel, Lori and Gauvain, Jean-Luc},
  booktitle={Proceedings of the 19th International Congress of Phonetic Sciences, Melbourne, Australia 2019},
  year={2019}
}

@inproceedings{gourav2025code,
  title={Code Mix TTS: An Approach to Infer Human Like Speech for Multi-Lingual Input Texts},
  author={Gourav, Vishal and Mankale, Phanindra},
  booktitle={Proc. Interspeech 2025},
  pages={2143--2144},
  year={2025}
}

@inproceedings{bhogale2026towards,
  title={Towards Orthographically-Informed Evaluation of Speech Recognition Systems for Indian Languages},
  author={Bhogale, Kaushal Santosh and Javed, Tahir and John, Greeshma Susan and Rathi, Dhruv and Padmanaban, Akshayasree and Parasa, Niharika and Khapra, Mitesh M},
  booktitle={ICASSP 2026-2026 IEEE International Conference on Acoustics, Speech and Signal Processing (ICASSP)},
  pages={17092--17096},
  year={2026},
  organization={IEEE}
}

@phdthesis{ahn2025investigating,
  author       = {Ahn, Emily Proch},
  title        = {Investigating the Corpus Phonetics Pipeline Applied to Diverse Speech Data},
  school       = {University of Washington},
  year         = {2025},
  address      = {Seattle, WA, USA},
  type         = {Ph.D. dissertation},
  url          = {https://www.proquest.com/openview/b2c6a1a23f8438923e916c82eecfedc9/1?pq-origsite=gscholar\&cbl=18750\&diss=y},
  note         = {ProQuest Dissertations \& Theses, publication number 32114404}
}

@inproceedings{ahn-etal-2025-automatic,
    title = "Automatic Phone Alignment of Code-switched {U}rum{--}{R}ussian Field Data",
    author = "Ahn, Emily  and
      Chodroff, Eleanor  and
      Levow, Gina-Anne",
    editor = "Le Ferrand, {\'E}ric  and
      Klyachko, Elena  and
      Postnikova, Anna  and
      Shavrina, Tatiana  and
      Serikov, Oleg  and
      Voloshina, Ekaterina  and
      Vylomova, Ekaterina",
    booktitle = "Proceedings of the Fourth Workshop on NLP Applications to Field Linguistics",
    month = aug,
    year = "2025",
    address = "Vienna, Austria",
    publisher = "Association for Computational Linguistics",
    url = "https://aclanthology.org/2025.fieldmatters-1.1/",
    pages = "1--14",
    ISBN = "979-8-89176-282-4"
}

@article{parasher1981indian,
  title={Indian {E}nglish: A sociolinguistic perspective},
  author={Parasher, SV},
  journal={ITL-International Journal of Applied Linguistics},
  volume={51},
  number={1},
  pages={59--70},
  year={1981},
  doi = {10.1075/itl.51.04par},
  publisher={John Benjamins}
}

@inproceedings{pandey2018phonetically,
  title={Phonetically balanced code-mixed speech corpus for {H}indi-{E}nglish automatic speech recognition},
  author={Pandey, Ayushi and Srivastava, Brij Mohan Lal and Kumar, Rohit and Nellore, Bhanu Teja and Teja, Kasi Sai and Gangashetty, Suryakanth V},
  booktitle={Proceedings of the Eleventh International Conference on Language Resources and Evaluation (LREC 2018)},
  year={2018}
}

@incollection{Tang_Bennett2019_bootstrapping_FA,
  author={Kevin Tang and Ryan Bennett},
  title={Unite and conquer: Bootstrapping forced alignment tools for closely-related minority languages ({M}ayan)},
  booktitle={Proceedings of the {19th International Congress of Phonetic Sciences, Melbourne, Australia 2019}},
  pages={1719-1723},
  year={2019},
  editor={Sasha Calhoun and Paola Escudero and Marija Tabain and Paul Warren},
  publisher={{Australasian Speech Science and Technology Association Inc.}},
  ADDRESS = {Canberra, Australia},
  url = {https://assta.org/proceedings/ICPhS2019/papers/ICPhS\_1768.pdf},
  isbn = {978-0-646-80069-1},
}

@inproceedings{buschmeier2013textgridtools,
  title = {{TextGridTools}: A TextGrid processing and analysis toolkit for {P}ython},
  author = {Buschmeier, Hendrik and Wlodarczak, Marcin},
  booktitle = {Conference proceedings of the 24th conference on electronic speech signal processing (ESSV 2013)},
  year = {2013}
}

\end{document}